\pdfoutput=1

\documentclass[11pt]{article}

\usepackage[final]{acl}

\usepackage{times}
\usepackage{latexsym}

\usepackage[T1]{fontenc}

\usepackage[utf8]{inputenc}

\usepackage{microtype}

\usepackage{inconsolata}

\usepackage{graphicx}
\usepackage{graphicx}
\usepackage{hyperref}
\usepackage{url}
\usepackage{array}
\usepackage{tabularx}
\usepackage{booktabs}
\usepackage{multicol}
\usepackage{multirow}
\usepackage{xcolor}
\usepackage{times}
\usepackage{float}
\usepackage{latexsym}
\usepackage{booktabs}
\usepackage{multirow}
\usepackage{amsmath}
\usepackage{float}
\usepackage{graphicx}
\usepackage{wrapfig}
\usepackage{subcaption}
\usepackage{subcaption}
\usepackage{placeins}
\usepackage[inkscapelatex=false]{svg}

\usepackage[T1]{fontenc}

\usepackage[utf8]{inputenc}

\usepackage{microtype}

\title{SceneGraMMi: Scene Graph-boosted Hybrid-fusion for Multi-Modal Misinformation Veracity Prediction}

\author{
  Swarang Joshi\thanks{Equal contribution.},\enspace Siddharth Mavani\footnotemark[1], \enspace Joel Alex,\enspace Arnav Negi,\enspace Rahul Mishra,\enspace Ponnurangam Kumaraguru\\
  International Institute of Information Technology, Hyderabad, India\\
  \texttt{swarang.joshi@research.iiit.ac.in}
}

\begin{document}
\maketitle
\begin{abstract}
Misinformation undermines individual knowledge and affects broader societal narratives. Despite growing interest in the research community in multi-modal misinformation detection, existing methods exhibit limitations in capturing semantic cues, key regions, and cross-modal similarities within multi-modal datasets. We propose SceneGraMMi, a \textbf{Scene} \textbf{G}raph-boosted Hybrid-fusion approach for \textbf{M}ulti-modal  \textbf{Mi}sinformation veracity prediction, which integrates scene graphs across different modalities to improve detection performance. Experimental results\footnote{All code, data, models, and detailed hyperparameters configurations will be publicly released for reproducibility.} across four benchmark datasets show that SceneGraMMi consistently outperforms state-of-the-art methods. In a comprehensive ablation study, we highlight the contribution of each component, while Shapley values are employed to examine the explainability of the model's decision-making process.
\end{abstract}

\begin{figure*}[h]
    \centering
    \includegraphics[width=1\textwidth]{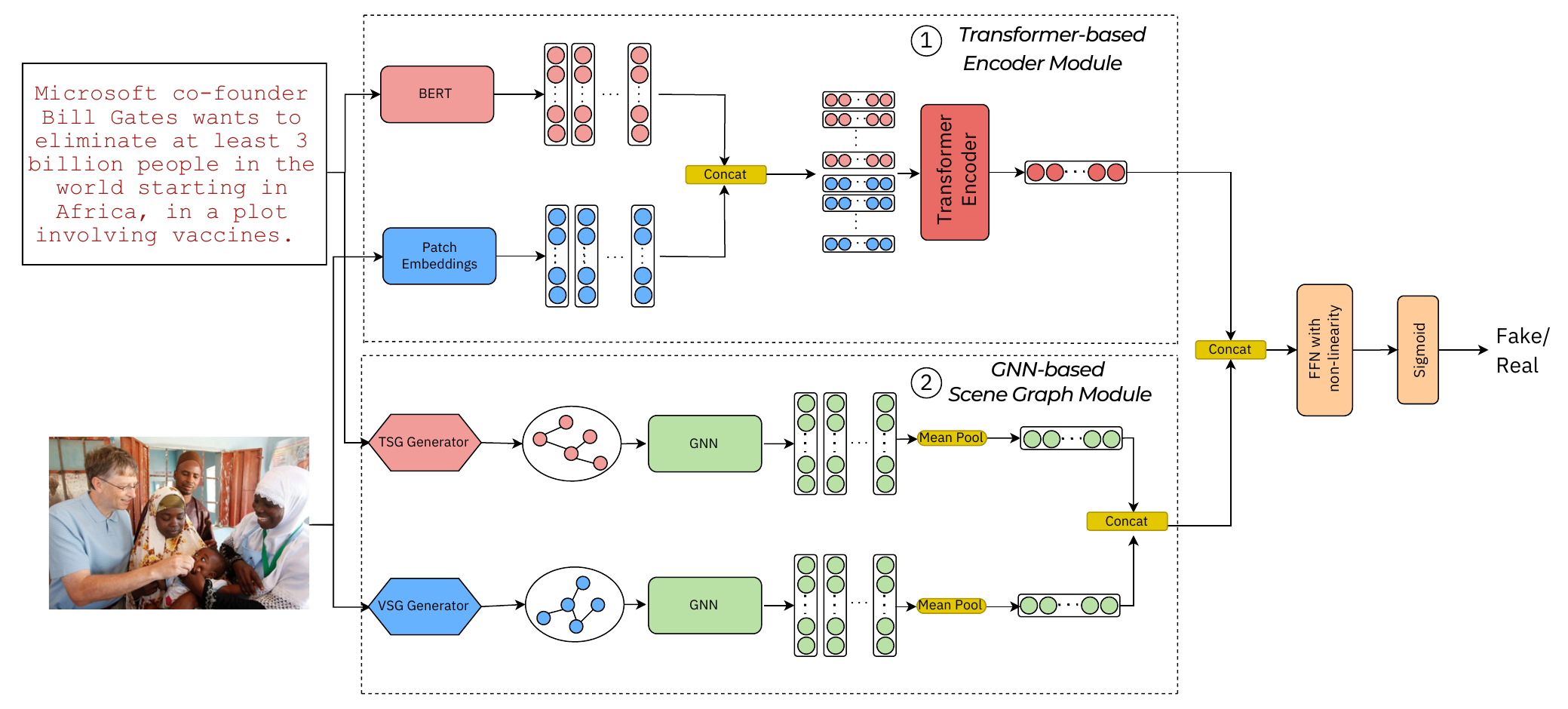}
    \caption{Architecture diagram of our multi-modal fake news detection model. Image-text pairs are passed as input to each module, and the final classification is done through a Feed Forward Network (FFN) to predict the final label: Fake or Real. The components of each module, namely the 1.) Transformer-based Encoder Module (TEM) and the 2.) GNN-based Scene Graph Module (GSGM), are enclosed in dotted boxes.}
    \label{fig:architecture}
\end{figure*}

\section{Introduction}
The advent of social media platforms like Twitter (now called X) and Weibo, replacing traditional mediums such as television and radio, has revolutionized the landscape of information \cite{Mitra2017APL, Wu2013AgendaSA}. This shift has not only made the process of information acquisition more efficient but also raised serious concerns. The ease of information spread has inadvertently led to the genesis and widespread propagation of misinformation and fake news, which are defined as false or misleading narratives disguised as news articles, intending to deceive users, resulting in harmful effects \cite{Allcott2017}.

Currently, established fact-checking platforms like PolitiFact\footnote{\url{https://www.politifact.com}} and Snopes\footnote{\url{https://www.snopes.com}} primarily focus on verifying the truthfulness of claims. However, the manual annotation of misinformation on these platforms requires significant human labor and is unable to meet the temporal demands of real-time fake news detection. This often results in the widespread proliferation of fake news before it can be identified. As a result, automated fake news detection systems have become a focal point of research in recent years \cite{Mishra_ictir19,10196320, Oshikawa2020, Shu2017, gupta2015tweetcredrealtimecredibilityassessment,Singhal2019}.

Existing fake news detection methods can be broadly categorized into two types: context-based methods and content-based methods \cite{Shu2017, ZhouZafarani2020}. Context-based methods model the task as a reasoning process, where external contexts, such as social contexts or textual contexts, are provided to help ascertain the veracity of a given claim \cite{hu2021compare, vlachos2014fact, wang2020fake, wu2021unified}. Social contexts refer to the user profile, social engagements, and propagation networks, \cite{9151025,li2022adadebunk, nguyen2020fang, silva2021propagation2vec} while textual contexts are usually additional factual sources retrieved from knowledge graphs \cite{vlachos2015identification} or fact-checking websites \cite{vlachos2014fact}. Although context-based methods offer preliminary interpretability due to the use of external resources, such contexts are not always available and are difficult to extract.

Consequently, an increasing number of approaches are adopting the content-based paradigm that focuses on using news content including text and images, and does not assume the availability of explicit external resources. These methods extract useful features from the news contents (or comments), such as text and image features/patterns, semantic information, and emotion signals, and feed them to neural models to predict the veracity \cite{zhang2021mining, przybyla2020capturing, ma2019detect, mishra-etal-2020-generating}. However, these content-based methods often fall short of capturing the complex relationships and dependencies between different modalities in the news item, which results in poor robustness and generalization capabilities of such models.

We propose a novel hybrid modality fusion model which employs a two-pronged approach: first, it uses late fusion to combine scene graphs generated from text and image pairs, providing a structured representation that captures relationships between various entities. Second, it incorporates early modality fusion by integrating sequences of text tokens and image patches through a transformer encoder. This enables a more nuanced and comprehensive understanding of both modalities by capturing the context and interdependencies between entities and objects present in both text and images.

\section{Related Work}
In recent years, the proliferation of fake news has become a pressing issue, significantly impacting public opinion, political discourses, and societal trust. Existing research on fake news \cite{Ruchansky2017CSIAH, ijcai2017p545, Wu2021} defines it as news that is intentionally fabricated and can be verified as false. Uni-modal and Multi-modal approaches for fake news detection have been extensively researched in recent years. 

\subsection{Uni-modal Fake News Detection}
Textual features extracted from news articles or social media posts have been extensively studied for fake news detection. Traditional approaches rely on handcrafted features such as statistical metrics (e.g., paragraph count, percentage of negative words, punctuation usage) and semantic features (e.g., writing styles, language styles) \cite{volkova-etal-2017-separating, potthast-etal-2018-stylometric, 10.1145/1963405.1963500, 10.1145/2823465.2823467, feng2012syntactic}. However, the reliance on handcrafted features introduces bias and design complexity. To overcome this limitation, recent studies have turned to deep learning techniques, including Recurrent Neural Networks (RNN), Convolutional Neural Networks (CNN), and Generative Adversarial Networks (GAN) \cite{Ma2016DetectingRF, yu2017convolutional, ma2019detect}, showcasing superior performance compared to traditional methods. User metadata, such as writing style, network connections, and user reactions, have also been explored for fake news detection \cite{potthast-etal-2018-stylometric} demonstrating the impact of author writing style on shaping readers' opinions. Image analysis through visual features such as clarity score and image count have been manually crafted for news verification \cite{Jin2016NewsVB, Shu2017} but limitations arise due to the simplistic nature of these features and their applicability to real images.

\subsection{Multi-modal Fake News Detection}
Multi-modal approaches to fake news detection have emerged as a promising avenue for enhancing detection accuracy by leveraging information from multiple modalities, including text, images, and social context.  \citet{Qi2021ImprovingFN} manually extracted entities and texts from images to complement textual content, addressing limitations associated with pre-trained feature extractors. \citet{zhang2021mining} introduced a novel dual emotion feature descriptor, demonstrating the efficacy of emotional cues in distinguishing fake from real news.\\
Frameworks that integrate textual, visual, and knowledge information to model semantic representations and enhance detection accuracy have also been developed such as KMGCN \cite{wang2020fake}, while \citet{MllerBudack2020MultimodalAF} quantified entity coherence between images and text for improved fake news detection. However, these methods require extensive computational resources and domain-specific knowledge for effective implementation. \citet{Singhal2019} proposed Spotfake  using VGG and BERT to extract features, which is then improved in Spotfake+ \cite{Singhal2020} for full-length article detection. SAFE \cite{Zhou2020} calculates relevance between textual and visual information, while MCAN \cite{Wu2021} utilized multiple co-attention layers to fuse multi-modal features.  \citet{Qian2021} employed hierarchical semantics of text and regional image representations, integrating them through co-attention layers for hierarchical feature fusion.

\begin{figure*}[h]
    \centering
    \includegraphics[width=.81\textwidth]{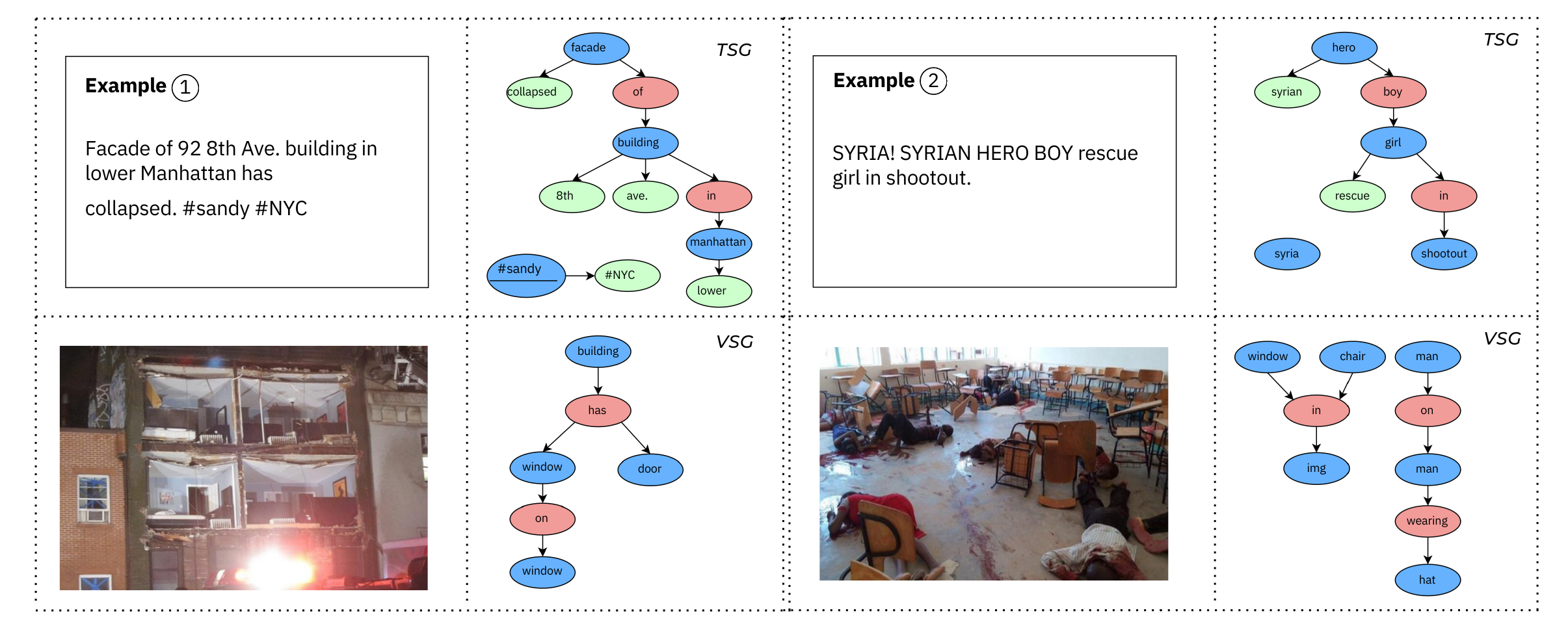}
    \caption{The diagram depicts the scene graphs of 2 samples in the Twitter dataset. Each sample has an input text and an input image along with the associated Scene Graphs. Object nodes are blue in color, attribute nodes are green in color and relationship nodes are red in color.}
    \label{fig:scene_graph_example}
\end{figure*}

\section{Methodology}

In this section, we describe the overall architecture of the SceneGraMMi, as depicted in Figure \ref{fig:architecture}.

\subsection{Model Outline}

We propose a novel hybrid modality fusion approach that combines content from both modalities through two stages: early fusion using a Transformer Encoder module and late fusion via a Graph Neural Network (GNN) \cite{4700287} applied to scene graphs from both modalities. The overall model comprises two main components: a Transformer-based Encoder Module (TEM) and a GNN-based Scene Graph Module (GSGM). The TEM processes a concatenated sequence of text tokens and image patches, while the GSGM takes as input the scene graphs generated from both the text and image. The learned representations from the TEM and GSGM are then merged and passed through a Feed Forward Netwrok (FFN) with a non-linearity to learn the final representation used for classification.

\subsubsection{Transformer-based Encoder Module (TEM)}
The encoder module in the multi-modal transformer consists of several key components that enable the model to process both textual and visual inputs effectively. At its core, the encoder module leverages the transformer architecture \cite{Vaswani2017}, known for its ability to handle sequential data such as text and images. The encoder module starts by tokenizing the input text using a BERT tokenizer \cite{Devlin2019}. These tokens are then passed through a pre-trained BERT model to obtain contextualized embeddings for each token. Each token is represented by a vector \(\mathbf{E}_{\text{text}} \in \mathbf{R}^{768} \). The embeddings capture the semantic meaning of the text, allowing the model to understand the context of the input.

Simultaneously, the input image is processed using a PatchEmbedding layer, which breaks down the image into smaller patches and flattens them into vectors. The size of each patch is 16x16 pixels, and each patch is flattened into a vector. The number of image patches depends on the dimensions of the input image, with larger images containing more patches. The final embedding dimension for each image patch is also represented by a vector \(\mathbf{E}_{\text{image}} \in \mathbf{R}^{768} \), consistent with the token embeddings from the text input. These patch embeddings are then concatenated with the textual embeddings, creating a combined representation of the input text and image (Figure \ref{fig:architecture}).

\begin{equation}
    \mathbf{E}_{\text{encoder}} = \mathbf{E}_{\text{text}} \oplus \mathbf{E}_{\text{image}}
\end{equation}

Combined embeddings are then passed through a series of encoder layers. Each encoder layer consists of a Multi-Head Attention mechanism, which allows the model to focus on different parts of the input data simultaneously. 

After the attention mechanism, the embeddings go through a FFN, which applies non-linear transformations to the embeddings, further capturing complex patterns in the data. The output of the FFN is then passed through a normalization layer and a dropout layer to improve the model's generalization ability and prevent overfitting.

\subsubsection{GNN-based Scene Graph Module (GSGM)}

Scene graphs  \cite{Li2017, Zhang2017, Yang2018} offer a structured representation of content, addressing contextual challenges often encountered in sequential text processing.

A scene graph $G$ (Figure \ref{fig:scene_graph_example}) is defined as a tuple $(V_O, V_A, V_R, E)$ where:
\begin{itemize}
    \item $V_O$ is the set of object nodes
    \item $V_A$ is the set of attribute nodes
    \item $V_R$ is the set of relationship nodes
    \item $V = V_O \cup V_A \cup V_R$ is the set of all nodes
    \item $E \subseteq (V_O \times V_R \times V_O) \cup (V_O \times V_A)$ is the set of directed edges
    \item Each node $v \in V$ is labeled with specific text $\text{label}(v)$
\end{itemize}

Within a scene graph, object, and attribute nodes are linked to other objects through pairwise relations. This structure inherently captures the semantic relationships within scene contexts, whether described in text or depicted in images. Pre-trained models are used for generating both the text scene graph (TSG) \cite{zhao2021semantic} and visual scene graph (VSG) \cite{wang2021reltr}. This data is then passed into separate GNNs to extract meaningful features from it. Each GNN consists of two Graph Convolutional Network (GCN) \cite{kipf2017semisupervisedclassificationgraphconvolutional} layers, which are specialized neural networks for processing graph-structured data. The GNN is initialized with an input dimension, which is the dimensionality of the node embeddings. 

After each GCN layer, a rectified linear unit (ReLU) \cite{agarap2019deeplearningusingrectified} activation function is applied to introduce non-linearity into the model. Finally, a global mean pooling operation is applied to aggregate node embeddings across the graph and obtain a single embedding for the entire graph.

Embeddings from the two GNNs are concatenated to create a single embedding representing both the visual and textual aspects of the scene graph data.

\begin{equation}
    \mathbf{E}_{\text{SG}} = \mathbf{E}_{\text{TSG}} \oplus \mathbf{E}_{\text{VSG}}
\end{equation}

The final embeddings are then passed on to the FFN to perform the final classification.

\subsubsection{Fake News Detector}

Embeddings from the encoder and scene graph module are concatenated to integrate the features of both modules, i.e., $E_{\text{encoder}}$ and $E_{\text{SG}}$.

\begin{equation}
    \mathbf{E}_{\text{final}} = \mathbf{E}_{\text{encoder}} \oplus \mathbf{E}_{\text{SG}}
\end{equation}

After concatenation, the embeddings are passed through the FFN with non-linearity.
Then we apply the integrated features to the Sigmoid function for task learning. The Sigmoid function gives us the prediction of the probability distribution. Given that the task is primarily a classification problem, we employ the binary cross-entropy loss as the objective function.

\section{Experiments}

\subsection{Datasets} 
For evaluating the performance\footnote{Details of the hyperparameters, and pre-trained modules and hardware used in our experiments can be found in Appendix \ref{expAppendix}.} of our model, we use 4 different datasets. These datasets are Weibo \cite{Jin2017}, Politifact $\And$ Gossipcop \cite{shu2017exploiting, Shu2017, shu2018fakenewsnet} and Twitter (MediaEval 15/16) \cite{boididou2018detection}. Weibo is a widely used Chinese dataset in fake news detection. The dataset is collected by \citet{Jin2017}. 

Politifact and Gossipcop datasets are two English datasets collected from the political and entertainment domains of the FakeNewsNet \cite{shu2018fakenewsnet} repository, respectively. The Twitter dataset was released for the MediaEval Verifying Multimedia Use task \cite{Chen2022} and is also a well-known multi-modal dataset for fake news detection. While these datasets contain both user-level and content-level data, we make use of just the content and curate a corpus of image-text pairs. The details for each dataset are shown in Table \ref{dataset_statistics}.

\begin{table}[h]
\centering
\renewcommand{\arraystretch}{0.90}
\scalebox{0.75}{
\begin{tabular}{l c c c c c c}
\toprule
 & \multicolumn{2}{c}{\textbf{Train}} & \multicolumn{2}{c}{\textbf{Test}} & \multicolumn{1}{c}{\textbf{}} \\
\cmidrule(lr){2-3} \cmidrule(lr){4-5}
\textbf{Dataset} & \textbf{Real} & \textbf{Fake} & \textbf{Real} & \textbf{Fake}  \\
\midrule
Twitter & 6357 & 9271 & 998 & 1230  \\
Weibo & 3783 & 3748 & 996 & 1000  \\
Politifact & 440 & 308 & 110 & 78  \\
Gossipcop & 5427 & 3936 & 1357 & 985 \\
\bottomrule
\end{tabular}
}
\caption{Statistics of the image-text pair distribution of Real and Fake news in the used datasets.}
\label{dataset_statistics}
\end{table}

\subsection{Baselines}
We compare our proposed model with the following models: \textbf{MMFN} \cite{Zhou2023} fuses fine-grained features with coarse-grained features encoded by the CLIP encoder and employs similarity-based weighting in uni-modal branches to adaptively adjust the use of multi-modal features; \textbf{CAFE} \cite{Chen2022} aligns heterogeneous uni-modal features into a shared space, estimates ambiguity between modalities, and captures cross-modal correlations; \textbf{HMCAN} \cite{Qian2021} utilizes a multi-modal contextual attention network to fuse these representations and a hierarchical encoding network to capture hierarchical semantics; \textbf{MCAN} \cite{Wu2021} learns inter-dependencies among modalities by extracting spatial and frequency-domain features from images, along with textual features, and fusing them through a deep co-attention model; \textbf{SpotFake+} \cite{Singhal2020} leverages transfer learning to capture semantic and contextual information from news articles and their associated images; \textbf{SAFE} \cite{Zhou2020} utilizes neural networks to extract textual and visual features from news articles and investigates the relationship between these features across modalities to detect fake news.

\subsubsection{Proposed Baselines For Scene Graphs}

In addition to benchmarking our model against these baselines, we conducted a series of experiments to evaluate the effectiveness of various techniques in generating and utilizing scene graphs from input text and images. Our investigation extended to assessing how different embeddings and model architectures influence performance. Our base model is composed of GNNs trained on text and image scene graphs. These scene graphs were generated from the input text and images. GloVe embeddings \cite{pennington-etal-2014-glove} were used to generate numerical representations of the words present in these scene graphs. 

We further experiment with the base model by incorporating node2vec (N2V) \cite{grover2016node2vecscalablefeaturelearning} graph embeddings. Additionally, we trained the model with a combination of N2V and GNN. The results of these experiments are presented in Table \ref{tab:n2v_experiments}. We also explore the use of a diffusion model to generate images from input text. Generated images were then fed solely into the encoder module. Unlike the original setup, these newly generated images were not processed by the scene graph module. Instead, separate patch embeddings were created for the generated images and subsequently concatenated with the embeddings from the input images and text. These concatenated embeddings were then fed into the transformer encoder, leading to the final classification. Results of this experiment are detailed in Table \ref{tab:n2v_experiments}.

We also experimented with the structure of the scene graphs. The base model uses two separate scene graphs for each modality (text and image) and employs two separate GNNs for each of them. We proposed a fusion of these two scene graphs, which we refer to as the Cross-Modal Scene Graph (CMSG). This graph was created using three different techniques:

1. \textbf{Dummy Node Technique (CMSG Type 1)}: We created a dummy node and linked it with every other node in each of the individual scene graphs. The node representation of this dummy node was then used for the final classification after applying Graph Convolution in the GNN.

2. \textbf{Exact Node Merging (CMSG Type 2)}: We merged nodes that were exactly the same across the two scene graphs.

3. \textbf{Similarity-Based Node Merging (CMSG Type 3)}: We merged nodes based on cosine similarity. This was done for a range of similarity thresholds.

Table \ref{tab:cmsg_table} summarizes the results for the above techniques. 

\begin{table*}[htb!]

\centering
\renewcommand{\arraystretch}{0.72}
\scalebox{0.75}{
\begin{tabular}{lcccccccc}
\toprule
 & \textbf{Method} & \textbf{Accuracy} & \multicolumn{3}{c}{\textbf{Fake News}} & \multicolumn{3}{c}{\textbf{Real News}} \\
\cmidrule(lr){4-6}
\cmidrule(lr){7-9}
 &  &  & \textbf{Precision} & \textbf{Recall} & \textbf{F1-score} & \textbf{Precision} & \textbf{Recall} & \textbf{F1-score} \\
\midrule
\multirow{8}{*}{Twitter} & SpotFake+ & 0.827 & 0.821 & 0.800 & 0.810 & 0.802 & 0.836 & 0.831 \\
 & CAFE & 0.806 & 0.807 & 0.799 & 0.803 & 0.805 & 0.813 & 0.809 \\
 & MCAN & 0.809 & 0.889 & 0.765 & 0.822 & 0.732 & 0.871 & 0.795 \\
 & SAFE & 0.856 & 0.865 & 0.870 & 0.860 & 0.865 & 0.862 & 0.867 \\ 
 & HMCAN & 0.897 & \textbf{0.971} & 0.801 & 0.878 & 0.853 & 0.979 & 0.912 \\
 & MMFN & 0.935 & 0.960 & 0.956 & 0.905 & 0.924 & \textbf{0.990} & 0.951 \\
 & \textbf{SceneGraMMi} & \textbf{0.945} & 0.961 & \textbf{0.959} & \textbf{0.941} & \textbf{0.940} & 0.942 & \textbf{0.953} \\
 \midrule
\multirow{7}{*}{Weibo} & SpotFake+ & 0.872 & 0.882 & 0.914 & 0.912 & 0.827 & 0.856 & 0.839 \\
 & CAFE & 0.840 & 0.855 & 0.830 & 0.842 & 0.825 & 0.851 & 0.837 \\
 & MCAN & 0.899 & 0.913 & 0.889 & 0.901 & 0.884 & 0.909 & 0.897 \\
 & SAFE & 0.880 & 0.892 & 0.882 & 0.879 & 0.885 & 0.876 & 0.872 \\
 & HMCAN & 0.885 & 0.920 & 0.845 & 0.881 & 0.856 & \textbf{0.926} & 0.890 \\
 & MMFN & 0.923 & 0.921 & 0.926 & 0.924 & 0.924 & 0.920 & 0.922 \\
 & \textbf{SceneGraMMi} & \textbf{0.928} & \textbf{0.929} & \textbf{0.931} & \textbf{0.925} & \textbf{0.927} & 0.921 & \textbf{0.923} \\
\midrule
\multirow{6}{*}{Politifact}  & SpotFake+ & 0.846 & 0.857 & 0.815 & 0.827 & 0.873 & 0.890 & 0.801 \\
& CAFE & 0.864 & 0.724 & 0.778 & 0.750 & 0.895 & 0.919 & 0.907 \\
& MCAN & 0.909 & 0.899 & 0.915 & 0.882 & 0.832 & 0.905 & 0.871 \\
& SAFE & 0.874 & 0.851 & 0.830 & 0.840 & 0.889 & 0.903 & 0.896 \\
& HMCAN & 0.907 & 0.870 & 0.862 & 0.877 & 0.906 & 0.940 & 0.923 \\
& MMFN & 0.923 & 0.912 & 0.909 & 0.915 & 0.937 & 0.941 & 0.931 \\
 & \textbf{SceneGraMMi} & \textbf{0.944} & \textbf{0.941} & \textbf{0.942} & \textbf{0.948} & \textbf{0.949} & \textbf{0.943} & \textbf{0.947} \\
\midrule
\multirow{6}{*}{Gossipcop} & SpotFake+ & 0.858 & 0.732 & 0.372 & 0.494 & 0.866 & 0.962 & 0.914 \\
  & CAFE & 0.867 & 0.732 & 0.490 & 0.587 & 0.887 & 0.957 & 0.921 \\
  & MCAN & 0.876 & 0.858 & 0.863 & 0.822 & 0.782 & 0.857 & 0.828 \\
& SAFE & 0.838 & 0.758 & 0.558 & 0.643 & 0.857 & 0.937 & 0.895 \\
& HMCAN & 0.869 & 0.852 & 0.860 & 0.852 & 0.874 & 0.883 & 0.861 \\
 & MMFN & \textbf{0.894} & 0.799 & 0.598 & 0.684 & \textbf{0.910} & \textbf{0.964} & \textbf{0.936} \\
 & \textbf{SceneGraMMi} & 0.871 & \textbf{0.870} & \textbf{0.869} & \textbf{0.865} & 0.862 & 0.861 & 0.870 \\
\bottomrule
\end{tabular}
}
\caption{Performance comparison between our model and state-of-the-art models on Politifact, Gossipcop, Twitter, and Weibo. The highest-performing model for each dataset has been bolded. Our model consistently outperforms or competes with the other benchmark models.}
\label{Results_Table}

\end{table*}

\section{Results}

In this section, we compare the performance of our model with baselines on the four datasets: Twitter, Weibo, Politifact, and Gossipcop. In Table \ref{Results_Table}, we report the empirical results of such a comparison according to the metrics considered.

For the Weibo dataset, which was translated to English from Chinese, our model outperformed the other models, achieving an accuracy of 92.8\%. 

Politifact, a dataset focused on the political domain, presents unique challenges due to its specialized content and relatively small size, containing just 936 samples. Despite these challenges, SceneGraMMi demonstrated strong performance on this dataset, achieving an accuracy of 94.4\%. 

For the Gossipcop dataset, which focuses on entertainment news, our model achieved competitive results. However, SceneGraMMi's performance was slightly below that of MMFN, a strong baseline model, which achieved the highest accuracy on this dataset. 

Nonetheless, SceneGraMMi's performance on the Gossipcop dataset demonstrates its effectiveness in handling diverse and subjective content. 
Overall, our model consistently outperforms or competes with state-of-the-art models across different datasets and domains, demonstrating its effectiveness and robustness in fake news detection. The superior performance can be attributed to the novel use of the encoder and scene graph modules, which effectively capture and utilize the semantic relationships within the text and images. 

\subsection{Scene Graph Baselines} Node2Vec (N2V) \cite{grover2016node2vecscalablefeaturelearning} captures the structural information of the graph, but it does not inherently understand the semantics or attributes associated with the nodes. When we use N2V alone, it might not perform as well as the base model because it only captures the structural relationships in the graph and misses out on the semantic information. On the other hand, GNNs are designed to handle graph-structured data and can capture both the graph structure and the node features. They propagate information through the graph and update the representation of each node based on its neighbors, thereby capturing local graph structure and node-level information. When we combine both N2V and GNNs, we are leveraging the strengths of both techniques which leads to an increase in performance as compared to just N2V.

The significant decrease in accuracy observed with the CMSG Type 1 can be attributed to the inherent information loss that occurs when a single dummy node is used to represent the features of both modalities simultaneously. Given the potentially large number of nodes in each scene graph, the use of a single representative node can lead to a substantial loss of information, thereby impacting the model’s accuracy. CMSG Type 2 also exhibits a decrease in accuracy, which can be ascribed to the limitations of merging nodes based on exact matches. For instance, consider a scenario where the TSG contains a node `man' and the VSG contains a node `boy', both referring to the same entity. In such a case, these nodes will not be merged due to the lack of an exact match, leading to an averaging of information and subsequent information loss. CMSG Type 3 allows for the adjustment of the similarity threshold, providing an opportunity to optimize results. As the threshold is reduced, an improvement in results is observed up to a certain point, beyond which the accuracy begins to decline again.

\subsection{Ablation Studies}

We present the results of our ablation studies to evaluate the contribution of each component in our model. Table \ref{ablations1} summarizes the results on the Politifact dataset, comparing the performance of the full SceneGraMMi model against variants where individual components— VSG, TSG, Image, or Text—are removed.

Excluding the VSG (Full SceneGraMMi w/o VSG) leads to a drop in accuracy from 94.4\% to 91.5\% and F1 Score from 92.6\% to 89.8\%, indicating its significant role in detecting fake news. Removing the TSG (Full SceneGraMMi w/o TSG) results in a more pronounced performance decline, with accuracy decreasing to 87.7\% and F1 Score to 83\%, underscoring the TSG's critical contribution. Both VSG and TSG are essential for capturing entity relationships in text and images, as demonstrated by these performance drops. When the image input is removed (Full SceneGraMMi w/o Image), accuracy decreases to 90.6\% and F1 Score to 88.9\%, showing that while images provide supplementary context, they are less critical than text. This aligns with the fact that fake news detection heavily relies on textual analysis, though images can sometimes offer contradictory or supportive context.

The most significant performance drop occurs when the text input is removed (Full SceneGraMMi w/o Text), with accuracy falling to 85\% and F1 Score to 83.3\%. This highlights the text's central role in disseminating news and its importance for detection tasks. The full model's superior performance demonstrates the efficacy of multi-modal learning, leveraging both text and images to enhance fake news detection, consistent with the multi-modal nature of misinformation.

\begin{table}[h]
\centering
\small
\scalebox{0.81}{
\begin{tabular}{l c c c c}
\toprule
\textbf{Model} & \textbf{Accuracy} & \textbf{F1 Score} & \textbf{Precision} & \textbf{Recall} \\
\midrule
CMSG Type 1 & 0.879 & 0.827 & 0.898 & 0.767 \\
CMSG Type 2 & 0.853 & 0.775 & 0.909 & 0.678 \\
\bottomrule
\end{tabular}
}
\caption{Performance of CMSG Type 1 and CMSG Type 2 Models. These models were trained on the Politifact dataset.}
\label{tab:cmsg_table}
\end{table}

\subsection{Error Analysis}

An analysis of the model's errors reveals distinct patterns in the misclassification of false negatives and false positives. False negatives reveal recurring issues where samples lack input images or contain minimal VSG data, hindering accurate classification. Sparse visual features and short texts with limited context lead to misclassification, emphasizing the need for improved handling of sparse data. In contrast, false positives often involve visually rich data with complex VSGs, where the abundance of entities and interconnections gives a misleading appearance of credibility.\begin{table}[h]
\centering
\begin{minipage}{0.48\textwidth}
\resizebox{\textwidth}{!}{
\begin{tabular}{l c c c c}
\toprule
\textbf{Model} & \textbf{Accuracy} & \textbf{F1 Score} & \textbf{Precision} & \textbf{Recall} \\
\midrule
SceneGraMMi\_N2V & 0.885 & 0.875 & 0.879 & 0.889 \\
SceneGraMMi\_N2V\_GNN & 0.925 & 0.894 & 0.921 & 0.869 \\
SceneGraMMi\_Diffusion & 0.927 & 0.918 &  0.964 & 0.876 \\
\bottomrule
\end{tabular}
}
\caption{The table summarizes the performance of our model SceneGraMMi when used along with N2V and N2V + GNN Models. The model was trained on the Politifact Dataset.}
\label{tab:n2v_experiments}
\end{minipage}
\end{table}

\begin{table}[h]
\centering
\begin{minipage}{0.48\textwidth}
\resizebox{\textwidth}{!}{
\begin{tabular}{l c c c c}
\toprule
\textbf{Model} & \textbf{Accuracy} & \textbf{F1 Score} & \textbf{Precision} & \textbf{Recall} \\
\midrule
SceneGraMMi & 0.944 & 0.926 & 0.962 & 0.893 \\
SceneGraMMi w/o VSG & 0.915 & 0.898 & 0.902 & 0.866 \\
SceneGraMMi w/o TSG & 0.877 & 0.830 & 0.894 & 0.860 \\
SceneGraMMi w/o Image & 0.906 & 0.889 & 0.922 & 0.857 \\
SceneGraMMi w/o Text & 0.850 & 0.803 & 0.845 & 0.834 \\
\bottomrule
\end{tabular}
}
\caption{Comparison of the proposed model and its ablations. SceneGraMMi w/o Image or Text is just with respect to the Encoder module, both modalities are still available in the Scene Graph module.}
\label{ablations1}
\end{minipage}
\end{table}These samples frequently contain detailed textual information, including structured content, further obscuring the fabricated nature of the content. The complexity of relationships in false positives enhances their perceived legitimacy, complicating detection.

\subsection{Explainability}

\begin{figure}[t]
  \centering
  \begin{subfigure}[b]{0.24\textwidth}
    \centering
    \includegraphics[width=\textwidth]{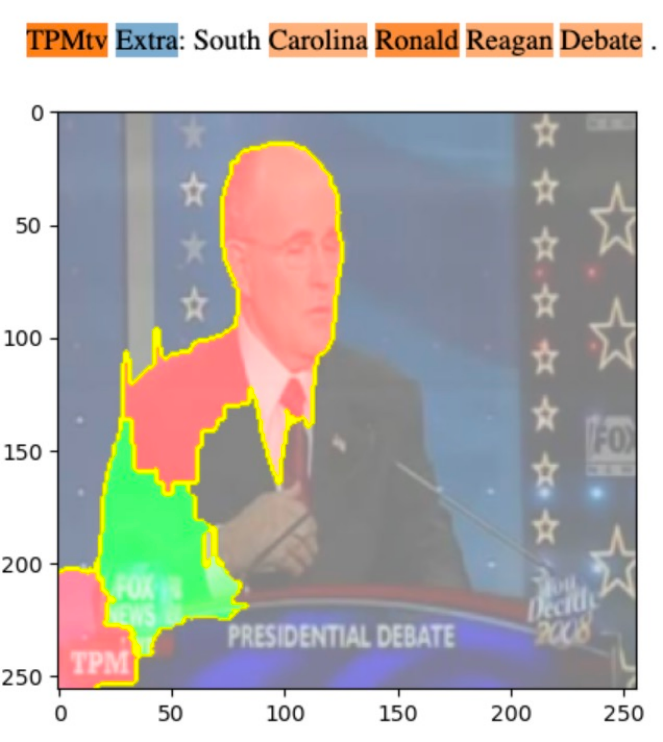}
    \caption{ }
    \label{fig:explain2a}
  \end{subfigure}
  \hfill
  \hspace{-3em}
  \begin{subfigure}[b]{0.24\textwidth}
    \centering
    \includegraphics[width=\textwidth]{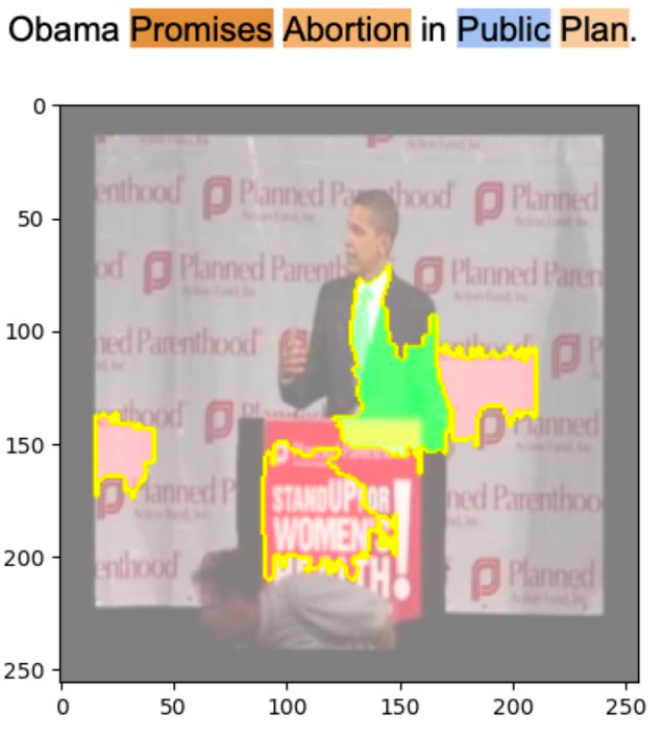}
    \caption{ }
    \label{fig:explain2b}
  \end{subfigure}
  \caption{Explainability of the model on example (a) and (b) which are selected from Politifact dataset. Each column consists of the image-text pair that is passed as input to the model. The highlighted portion of each modality is what the model uses to classify the sample as Fake or Real. Shapley values are utilised for highlighting model focus.}
  \label{fig:explain}
\end{figure}

In Figure \ref{fig:explain2a}, an image of a speech at a presidential debate is observed. The image consists of a man speaking in a presidential debate. The model is able to focus and identify the man, the fact that he is speaking on FOX News and also TPM. This is further reinforced by the content available in the text. The text also focuses on`TPMtv', which is in line with the image, and also the name of a male speaker, `Ronald Reagan'. The data highlighted in orange/red is what the model uses in favor of classifying as Not Fake/Real and blue/green is used in favor of classifying as Fake.

Similarly, in Figure \ref{fig:explain2b}, an image of a public speech is studied. The model is again able to identify and focus on the person speaking as well as the relevant text which is `Stand Up for Women'. The data highlighted in the text is also `Abortion' which is relevant to the women's rights issue being discussed in the data point. This shows that our model is able to identify and abstract away important information from both modalities and then perform the necessary classification.

\section{Discussion and Conclusions}

This study introduces a novel approach for fake news detection using multi-modal data, combining text and images through a transformer and scene graph module. The transformer processes each modality separately, while the scene graph captures semantic relationships within the data. This synergy allows the model to more accurately detect misinformation. Our approach consistently outperforms or competes with state-of-the-art methods across diverse datasets, achieving high accuracy, precision, recall, and F1-scores. The results underscore the model's robustness and adaptability to different topics and languages. By effectively integrating multi-modal data, our model provides a more comprehensive solution for fake news detection, contributing to ongoing efforts in combating misinformation. This work not only demonstrates the effectiveness of using multi-modal approaches but also offers a strong foundation for future research and applications in the field of fake news detection.

In terms of future work, one direction is to enhance the model's scalability and efficiency, possibly by exploring more lightweight architectures or techniques to reduce computational complexity. Additionally, exploring the use of other modalities, such as audio or video, could further enhance the model's capabilities in detecting fake news across a wider range of media types. Similarly, extending the model to languages other than English is a promising direction. This could involve adapting the model to handle multilingual datasets or leveraging cross-lingual transfer learning techniques. Additionally, fine-tuning language-specific components could significantly enhance its effectiveness across diverse linguistic contexts. Overall, while SceneGraMMi shows promise in fake news detection, there are still opportunities for further research and development to enhance its performance and applicability.


\section*{Limitations} While SceneGraMMi demonstrates strong performance across multiple datasets and domains, it also faces certain challenges and has areas for improvement. We identify the following limitations of our study.\\

\textbf{Monolingual constraint} SceneGraMMi is currently designed and trained for English language, providing strong performance in that context.\\

\textbf{Computational intensity} The early and late fusion techniques employed in our architecture, while effective, incur significant computational costs. Thus the ablation studies were performed on Politifact. \\

\textbf{Potential biases}  Model's reliance on pre-trained embeddings and scene graph generation may lead to biases present in these pre-trained models, affecting the model's performance on certain types of content or domains.
\\

\nocite{Allcott2017, Zannettou2019, Ruchansky2017CSIAH, ijcai2017p545, volkova-etal-2017-separating, potthast-etal-2018-stylometric, ApukeOmar2021, DiDomenico2021, Oshikawa2020, Shu2017, ZhouZafarani2020, hu2021compare, li2022adadebunk, nguyen2020fang, silva2021propagation2vec, vlachos2014fact, wang2020fake, wu2021evidence, wu2021unified, vlachos2015identification, lu2020gcan, ma2019detect, petratos2021misinformation, przybyla2020capturing, shu2019defend, talwar2019why, zhang2021mining, Vaswani2017, Devlin2019, Li2017, Yang2018, Zhang2017, wang2021reltr, zhao2021semantic, shu2017exploiting, shu2018fakenewsnet, boididou2018detection, Jin2017, vo2020facts, Singhal2019, Singhal2020, Chen2022, Wu2021, Singhal2022, Qian2021, Zhou2023, Allein2021, Zhou2020}

\bibliography{acl_latex}

\begin{thebibliography}{70}
\expandafter\ifx\csname natexlab\endcsname\relax\def\natexlab#1{#1}\fi

\bibitem[{Agarap(2019)}]{agarap2019deeplearningusingrectified}
Abien~Fred Agarap. 2019.
\newblock \href {http://arxiv.org/abs/1803.08375} {Deep learning using rectified linear units (relu)}.

\bibitem[{Allcott and Gentzkow(2017)}]{Allcott2017}
Hunt Allcott and Matthew Gentzkow. 2017.
\newblock Social media and fake news in the 2016 election.
\newblock \emph{Journal of economic perspectives}, 31(2):211--236.

\bibitem[{Allein et~al.(2021)Allein, Moens, and Perrotta}]{Allein2021}
Liesbeth Allein, Marie-Francine Moens, and Domenico Perrotta. 2021.
\newblock Like article, like audience: Enforcing multimodal correlations for disinformation detection.
\newblock \emph{arXiv preprint arXiv:2108.13892}.

\bibitem[{Anderson et~al.(2016)Anderson, Fernando, Johnson, and Gould}]{spice2016}
Peter Anderson, Basura Fernando, Mark Johnson, and Stephen Gould. 2016.
\newblock Spice: Semantic propositional image caption evaluation.
\newblock In \emph{ECCV}.

\bibitem[{Apuke and Omar(2021)}]{ApukeOmar2021}
Oberiri~Destiny Apuke and Bahiyah Omar. 2021.
\newblock \href {https://doi.org/10.1016/j.tele.2020.101475} {Fake news and covid-19: modelling the predictors of fake news sharing among social media users}.
\newblock \emph{Telematics and Informatics}, 56:101475.

\bibitem[{Castillo et~al.(2011)Castillo, Mendoza, and Poblete}]{10.1145/1963405.1963500}
Carlos Castillo, Marcelo Mendoza, and Barbara Poblete. 2011.
\newblock \href {https://doi.org/10.1145/1963405.1963500} {Information credibility on twitter}.
\newblock In \emph{Proceedings of the 20th International Conference on World Wide Web}, WWW '11, page 675–684, New York, NY, USA. Association for Computing Machinery.

\bibitem[{Chen et~al.(2015)Chen, Conroy, and Rubin}]{10.1145/2823465.2823467}
Yimin Chen, Niall~J. Conroy, and Victoria~L. Rubin. 2015.
\newblock \href {https://doi.org/10.1145/2823465.2823467} {Misleading online content: Recognizing clickbait as "false news"}.
\newblock In \emph{Proceedings of the 2015 ACM on Workshop on Multimodal Deception Detection}, WMDD '15, page 15–19, New York, NY, USA. Association for Computing Machinery.

\bibitem[{Chen et~al.(2022)Chen, Li, Zhang, Sui, Lv, Tun, and Shang}]{Chen2022}
Yixuan Chen, Dongsheng Li, Peng Zhang, Jie Sui, Qin Lv, Lu~Tun, and Li~Shang. 2022.
\newblock Cross-modal ambiguity learning for multimodal fake news detection.
\newblock In \emph{Proceedings of The Web Conference}, pages 2897--2905.

\bibitem[{Cong et~al.(2023)Cong, Yang, and Rosenhahn}]{cong2023reltr}
Yuren Cong, Michael~Ying Yang, and Bodo Rosenhahn. 2023.
\newblock Reltr: Relation transformer for scene graph generation.
\newblock \emph{IEEE Transactions on Pattern Analysis and Machine Intelligence}.

\bibitem[{Detection and visualization of misleading content~on Twitter(2018)}]{boididou2018detection}
Detection and visualization of misleading content~on Twitter. 2018.
\newblock \href {https://doi.org/10.1007/s13735-017-0143-x} {Boididou, christina and papadopoulos, symeon and zampoglou, markos and apostolidis, lazaros and papadopoulou, olga and kompatsiaris, yiannis}.
\newblock \emph{International Journal of Multimedia Information Retrieval}, 7(1):71--86.

\bibitem[{Devlin et~al.(2019)Devlin, Chang, Lee, and Toutanova}]{Devlin2019}
Jacob Devlin, Ming-Wei Chang, Kenton Lee, and Kristina Toutanova. 2019.
\newblock Bert: Pre-training of deep bidirectional transformers for language understanding.
\newblock \emph{arXiv preprint arXiv:1810.04805}.

\bibitem[{Di~Domenico et~al.(2021)Di~Domenico, Sit, Ishizaka, and Nunan}]{DiDomenico2021}
Giandomenico Di~Domenico, Jason Sit, Alessio Ishizaka, and Daniel Nunan. 2021.
\newblock \href {https://doi.org/10.1016/j.jbusres.2020.11.037} {Fake news, social media and marketing: A systematic review}.
\newblock \emph{Journal of Business Research}, 124:329--341.

\bibitem[{Feng et~al.(2012)Feng, Banerjee, and Choi}]{feng2012syntactic}
Song Feng, Ritwik Banerjee, and Yejin Choi. 2012.
\newblock Syntactic stylometry for deception detection.
\newblock In \emph{Proceedings of the 50th Annual Meeting of the Association for Computational Linguistics (Volume 2: Short Papers)}, pages 171--175.

\bibitem[{Grover and Leskovec(2016)}]{grover2016node2vecscalablefeaturelearning}
Aditya Grover and Jure Leskovec. 2016.
\newblock \href {http://arxiv.org/abs/1607.00653} {node2vec: Scalable feature learning for networks}.

\bibitem[{Gupta et~al.(2015)Gupta, Kumaraguru, Castillo, and Meier}]{gupta2015tweetcredrealtimecredibilityassessment}
Aditi Gupta, Ponnurangam Kumaraguru, Carlos Castillo, and Patrick Meier. 2015.
\newblock \href {http://arxiv.org/abs/1405.5490} {Tweetcred: Real-time credibility assessment of content on twitter}.

\bibitem[{Hu et~al.(2021)Hu, Yang, Zhang, Zhong, Tang, Shi, Duan, and Zhou}]{hu2021compare}
Linmei Hu, Tianchi Yang, Luhao Zhang, Wanjun Zhong, Duyu Tang, Chuan Shi, Nan Duan, and Ming Zhou. 2021.
\newblock Compare to the knowledge: Graph neural fake news detection with external knowledge.
\newblock In \emph{Proceedings of the 59th Annual Meeting of the Association for Computational Linguistics and the 11th International Joint Conference on Natural Language Processing (Volume 1: Long Papers)}, pages 754--763.

\bibitem[{Jin et~al.(2017)Jin, Cao, Guo, Zhang, and Luo}]{Jin2017}
Z.~Jin, J.~Cao, H.~Guo, Y.~Zhang, and J.~Luo. 2017.
\newblock \href {https://doi.org/10.1145/3123266.3123454} {Multimodal fusion with recurrent neural networks for rumor detection on microblogs}.
\newblock In \emph{Proceedings of the 2017 ACM on Multimedia Conference, MM 2017}, pages 795--816, Mountain View, CA, USA. ACM.

\bibitem[{Jin et~al.(2016)Jin, Cao, Zhang, and Luo}]{Jin2016NewsVB}
Zhiwei Jin, Juan Cao, Yongdong Zhang, and Jiebo Luo. 2016.
\newblock \href {https://api.semanticscholar.org/CorpusID:40056496} {News verification by exploiting conflicting social viewpoints in microblogs}.
\newblock In \emph{AAAI Conference on Artificial Intelligence}.

\bibitem[{Kipf and Welling(2017)}]{kipf2017semisupervisedclassificationgraphconvolutional}
Thomas~N. Kipf and Max Welling. 2017.
\newblock \href {http://arxiv.org/abs/1609.02907} {Semi-supervised classification with graph convolutional networks}.

\bibitem[{Kozik et~al.(2023)Kozik, Pawlicka, Pawlicki, Choraś, Mazurczyk, and Cabaj}]{10196320}
Rafał Kozik, Aleksandra Pawlicka, Marek Pawlicki, Michał Choraś, Wojciech Mazurczyk, and Krzysztof Cabaj. 2023.
\newblock \href {https://doi.org/10.1109/TCSS.2023.3296627} {A meta-analysis of state-of-the-art automated fake news detection methods}.
\newblock \emph{IEEE Transactions on Computational Social Systems}, pages 1--11.

\bibitem[{Li et~al.(2022)Li, Guo, Ren, and Yu}]{li2022adadebunk}
Ke~Li, Bin Guo, Siyuan Ren, and Zhiwen Yu. 2022.
\newblock Adadebunk: An efficient and reliable deep state space model for adaptive fake news early detection.
\newblock In \emph{Proceedings of the 31st ACM International Conference on Information \& Knowledge Management}, pages 1156--1165.

\bibitem[{Li et~al.(2017)Li, Ouyang, Bors, and Wang}]{Li2017}
Yikang Li, Wanli Ouyang, Adrian Bors, and Liang Wang. 2017.
\newblock Visual relationship detection with language priors.
\newblock In \emph{Proceedings of the IEEE Conference on Computer Vision and Pattern Recognition (CVPR)}.

\bibitem[{Lu and Li(2020)}]{lu2020gcan}
Yi-Ju Lu and Cheng-Te Li. 2020.
\newblock Gcan: Graph-aware co-attention networks for explainable fake news detection on social media.
\newblock In \emph{Proceedings of the 58th Annual Meeting of the Association for Computational Linguistics}, pages 505--514.

\bibitem[{Ma et~al.(2016)Ma, Gao, Mitra, Kwon, Jansen, Wong, and Cha}]{Ma2016DetectingRF}
Jing Ma, Wei Gao, Prasenjit Mitra, Sejeong Kwon, Bernard~Jim Jansen, Kam-Fai Wong, and M.~Cha. 2016.
\newblock \href {https://api.semanticscholar.org/CorpusID:16985095} {Detecting rumors from microblogs with recurrent neural networks}.
\newblock In \emph{International Joint Conference on Artificial Intelligence}.

\bibitem[{Ma et~al.(2019)Ma, Gao, and Wong}]{ma2019detect}
Jing Ma, Wei Gao, and Kam-Fai Wong. 2019.
\newblock Detect rumors on twitter by promoting information campaigns with generative adversarial learning.
\newblock In \emph{The World Wide Web Conference}, pages 3049--3055.

\bibitem[{Mishra(2020)}]{9151025}
Rahul Mishra. 2020.
\newblock \href {https://doi.org/10.1109/CVPRW50498.2020.00334} {Fake news detection using higher-order user to user mutual-attention progression in propagation paths}.
\newblock In \emph{2020 IEEE/CVF Conference on Computer Vision and Pattern Recognition Workshops (CVPRW)}, pages 2775--2783.

\bibitem[{Mishra et~al.(2020)Mishra, Gupta, and Leippold}]{mishra-etal-2020-generating}
Rahul Mishra, Dhruv Gupta, and Markus Leippold. 2020.
\newblock \href {https://doi.org/10.18653/v1/2020.wnut-1.12} {Generating fact checking summaries for web claims}.
\newblock In \emph{Proceedings of the Sixth Workshop on Noisy User-generated Text (W-NUT 2020)}, pages 81--90, Online. Association for Computational Linguistics.

\bibitem[{Mishra and Setty(2019)}]{Mishra_ictir19}
Rahul Mishra and Vinay Setty. 2019.
\newblock \href {https://doi.org/10.1145/3341981.3344229} {Sadhan: Hierarchical attention networks to learn latent aspect embeddings for fake news detection}.
\newblock ICTIR '19, New York, NY, USA. Association for Computing Machinery.

\bibitem[{Mitra et~al.(2017)Mitra, Wright, and Gilbert}]{Mitra2017APL}
Tanushree Mitra, Graham~P. Wright, and Eric Gilbert. 2017.
\newblock \href {https://api.semanticscholar.org/CorpusID:2037448} {A parsimonious language model of social media credibility across disparate events}.
\newblock \emph{Proceedings of the 2017 ACM Conference on Computer Supported Cooperative Work and Social Computing}.

\bibitem[{M{\"u}ller-Budack et~al.(2020)M{\"u}ller-Budack, Theiner, Diering, Idahl, and Ewerth}]{MllerBudack2020MultimodalAF}
Eric M{\"u}ller-Budack, Jonas Theiner, Sebastian Diering, Maximilian Idahl, and Ralph Ewerth. 2020.
\newblock \href {https://api.semanticscholar.org/CorpusID:214612150} {Multimodal analytics for real-world news using measures of cross-modal entity consistency}.
\newblock \emph{Proceedings of the 2020 International Conference on Multimedia Retrieval}.

\bibitem[{Nguyen et~al.(2020)Nguyen, Sugiyama, Nakov, and Kan}]{nguyen2020fang}
Van-Hoang Nguyen, Kazunari Sugiyama, Preslav Nakov, and Min-Yen Kan. 2020.
\newblock Fang: Leveraging social context for fake news detection using graph representation.
\newblock In \emph{Proceedings of the 29th ACM international conference on information \& knowledge management}, pages 1165--1174.

\bibitem[{Oshikawa et~al.(2020)Oshikawa, Qian, and Wang}]{Oshikawa2020}
Ray Oshikawa, Jing Qian, and William~Yang Wang. 2020.
\newblock \href {https://arxiv.org/abs/1811.00770} {A survey on natural language processing for fake news detection}.
\newblock In \emph{Proceedings of the 12th Language Resources and Evaluation Conference}, pages 6086--6093. European Language Resources Association (ELRA).

\bibitem[{Pennington et~al.(2014)Pennington, Socher, and Manning}]{pennington-etal-2014-glove}
Jeffrey Pennington, Richard Socher, and Christopher Manning. 2014.
\newblock \href {https://doi.org/10.3115/v1/D14-1162} {{G}lo{V}e: Global vectors for word representation}.
\newblock In \emph{Proceedings of the 2014 Conference on Empirical Methods in Natural Language Processing ({EMNLP})}, pages 1532--1543, Doha, Qatar. Association for Computational Linguistics.

\bibitem[{Petratos(2021)}]{petratos2021misinformation}
Pythagoras~N Petratos. 2021.
\newblock Misinformation, disinformation, and fake news: Cyber risks to business.
\newblock \emph{Business Horizons}, 64(6):763--774.

\bibitem[{Potthast et~al.(2018)Potthast, Kiesel, Reinartz, Bevendorff, and Stein}]{potthast-etal-2018-stylometric}
Martin Potthast, Johannes Kiesel, Kevin Reinartz, Janek Bevendorff, and Benno Stein. 2018.
\newblock \href {https://doi.org/10.18653/v1/P18-1022} {A stylometric inquiry into hyperpartisan and fake news}.
\newblock In \emph{Proceedings of the 56th Annual Meeting of the Association for Computational Linguistics (Volume 1: Long Papers)}, pages 231--240, Melbourne, Australia. Association for Computational Linguistics.

\bibitem[{Przybyla(2020)}]{przybyla2020capturing}
Piotr Przybyla. 2020.
\newblock Capturing the style of fake news.
\newblock In \emph{Proceedings of the AAAI Conference on Artificial Intelligence}, volume~34, pages 490--497.

\bibitem[{Qi et~al.(2021)Qi, Cao, Li, Liu, Sheng, Mi, He, Lv, Guo, and Yu}]{Qi2021ImprovingFN}
Peng Qi, Juan Cao, Xirong Li, Huan Liu, Qiang Sheng, Xiaoyue Mi, Qin He, Yongbiao Lv, Chenyang Guo, and Yingchao Yu. 2021.
\newblock \href {https://api.semanticscholar.org/CorpusID:237278121} {Improving fake news detection by using an entity-enhanced framework to fuse diverse multimodal clues}.
\newblock \emph{Proceedings of the 29th ACM International Conference on Multimedia}.

\bibitem[{Qian et~al.(2021)Qian, Wang, Hu, Fang, and Xu}]{Qian2021}
Shengsheng Qian, Jinguang Wang, Jun Hu, Quan Fang, and Changsheng Xu. 2021.
\newblock Hierarchical multi-modal contextual attention network for fake news detection.
\newblock In \emph{Proceedings of the International ACM SIGIR Conference on Research and Development in Information Retrieval}, pages 153--162.

\bibitem[{Ruchansky et~al.(2017)Ruchansky, Seo, and Liu}]{Ruchansky2017CSIAH}
Natali Ruchansky, Sungyong Seo, and Yan Liu. 2017.
\newblock \href {https://api.semanticscholar.org/CorpusID:5156607} {Csi: A hybrid deep model for fake news detection}.
\newblock \emph{Proceedings of the 2017 ACM on Conference on Information and Knowledge Management}.

\bibitem[{Scarselli et~al.(2009)Scarselli, Gori, Tsoi, Hagenbuchner, and Monfardini}]{4700287}
Franco Scarselli, Marco Gori, Ah~Chung Tsoi, Markus Hagenbuchner, and Gabriele Monfardini. 2009.
\newblock \href {https://doi.org/10.1109/TNN.2008.2005605} {The graph neural network model}.
\newblock \emph{IEEE Transactions on Neural Networks}, 20(1):61--80.

\bibitem[{Shu et~al.(2019)Shu, Cui, Wang, Lee, and Liu}]{shu2019defend}
Kai Shu, Limeng Cui, Suhang Wang, Dongwon Lee, and Huan Liu. 2019.
\newblock defend: Explainable fake news detection.
\newblock In \emph{Proceedings of the 25th ACM SIGKDD International Conference on Knowledge Discovery \& Data Mining}, pages 395--405.

\bibitem[{Shu et~al.(2018)Shu, Mahudeswaran, Wang, Lee, and Liu}]{shu2018fakenewsnet}
Kai Shu, Deepak Mahudeswaran, Suhang Wang, Dongwon Lee, and Huan Liu. 2018.
\newblock Fakenewsnet: A data repository with news content, social context and dynamic information for studying fake news on social media.
\newblock \emph{arXiv preprint arXiv:1809.01286}.

\bibitem[{Shu et~al.(2017{\natexlab{a}})Shu, Sliva, Wang, Tang, and Liu}]{Shu2017}
Kai Shu, Amy Sliva, Suhang Wang, Jiliang Tang, and Huan Liu. 2017{\natexlab{a}}.
\newblock \href {https://arxiv.org/pdf/1708.01967.pdf} {Fake news detection on social media: A data mining perspective}.
\newblock \emph{ACM SIGKDD Explorations Newsletter}, 19(1):22--36.

\bibitem[{Shu et~al.(2017{\natexlab{b}})Shu, Wang, and Liu}]{shu2017exploiting}
Kai Shu, Suhang Wang, and Huan Liu. 2017{\natexlab{b}}.
\newblock Exploiting tri-relationship for fake news detection.
\newblock \emph{arXiv preprint arXiv:1712.07709}.

\bibitem[{Silva et~al.(2021)Silva, Han, Luo, Karunasekera, and Leckie}]{silva2021propagation2vec}
Amila Silva, Yi~Han, Ling Luo, Shanika Karunasekera, and Christopher Leckie. 2021.
\newblock Propagation2vec: Embedding partial propagation networks for explainable fake news early detection.
\newblock \emph{Information Processing \& Management}, 58(5):102618.

\bibitem[{Singhal et~al.(2020)Singhal, Kabra, Sharma, Shah, Chakraborty, and Kumaraguru}]{Singhal2020}
Shivangi Singhal, Anubha Kabra, Mohit Sharma, Rajiv~Ratn Shah, Tanmoy Chakraborty, and Ponnurangam Kumaraguru. 2020.
\newblock Spotfake+: A multimodal framework for fake news detection via transfer learning (student abstract).
\newblock In \emph{Proceedings of the AAAI Conference on Artificial Intelligence}, volume~34, pages 13915--13916.

\bibitem[{Singhal et~al.(2019)Singhal, Shah, Chakraborty, Kumaraguru, and Satoh}]{Singhal2019}
Shivangi Singhal, Rajiv~Ratn Shah, Tanmoy Chakraborty, Ponnurangam Kumaraguru, and Shin'ichi Satoh. 2019.
\newblock Spotfake: A multi-modal framework for fake news detection.
\newblock In \emph{Proceedings of the IEEE International Conference on Multimedia Big Data}, pages 39--47.

\bibitem[{Talwar et~al.(2019)Talwar, Dhir, Kaur, Zafar, and Alrasheedy}]{talwar2019why}
Shalini Talwar, Amandeep Dhir, Puneet Kaur, Nida Zafar, and Melfi Alrasheedy. 2019.
\newblock Why do people share fake news? associations between the dark side of social media use and fake news sharing behavior.
\newblock \emph{Journal of Retailing and Consumer Services}, 51:72--82.

\bibitem[{Vaswani et~al.(2017)Vaswani, Shazeer, Parmar, Uszkoreit, Jones, Gomez, Kaiser, and Polosukhin}]{Vaswani2017}
Ashish Vaswani, Noam Shazeer, Niki Parmar, Jakob Uszkoreit, Llion Jones, Aidan~N. Gomez, Lukasz Kaiser, and Illia Polosukhin. 2017.
\newblock \href {http://papers.nips.cc/paper/7181-attention-is-all-you-need.pdf} {Attention is all you need}.
\newblock In \emph{Advances in Neural Information Processing Systems 30 (NIPS 2017)}, pages 5998--6008. Curran Associates, Inc.

\bibitem[{Vlachos and Riedel(2014)}]{vlachos2014fact}
Andreas Vlachos and Sebastian Riedel. 2014.
\newblock Fact checking: Task definition and dataset construction.
\newblock In \emph{Proceedings of the ACL 2014 workshop on language technologies and computational social science}, pages 18--22.

\bibitem[{Vlachos and Riedel(2015)}]{vlachos2015identification}
Andreas Vlachos and Sebastian Riedel. 2015.
\newblock Identification and verification of simple claims about statistical properties.
\newblock In \emph{Proceedings of the 2015 Conference on Empirical Methods in Natural Language Processing}, pages 2596--2601. Association for Computational Linguistics.

\bibitem[{Vo and Lee(2020)}]{vo2020facts}
Nguyen Vo and Kyumin Lee. 2020.
\newblock Where are the facts? searching for fact-checked information to alleviate the spread of fake news.
\newblock In \emph{Proceedings of the 2020 Conference on Empirical Methods in Natural Language Processing (EMNLP 2020)}.

\bibitem[{Volkova et~al.(2017)Volkova, Shaffer, Jang, and Hodas}]{volkova-etal-2017-separating}
Svitlana Volkova, Kyle Shaffer, Jin~Yea Jang, and Nathan Hodas. 2017.
\newblock \href {https://doi.org/10.18653/v1/P17-2102} {Separating facts from fiction: Linguistic models to classify suspicious and trusted news posts on {T}witter}.
\newblock In \emph{Proceedings of the 55th Annual Meeting of the Association for Computational Linguistics (Volume 2: Short Papers)}, pages 647--653, Vancouver, Canada. Association for Computational Linguistics.

\bibitem[{Wang et~al.(2021)Wang, Li, Zhao, Yang, and Zhang}]{wang2021reltr}
Xindi Wang, Wenbin Li, Boxin Zhao, Shuang Yang, and Haohong Zhang. 2021.
\newblock Reltr: Relation transformer for scene graph generation.
\newblock \emph{arXiv preprint arXiv:2106.01889}.

\bibitem[{Wang et~al.(2020)Wang, Qian, Hu, Fang, and Xu}]{wang2020fake}
Youze Wang, Shengsheng Qian, Jun Hu, Quan Fang, and Changsheng Xu. 2020.
\newblock Fake news detection via knowledge-driven multimodal graph convolutional networks.
\newblock In \emph{Proceedings of the 2020 international conference on multimedia retrieval}, pages 540--547.

\bibitem[{Wolf et~al.(2020)Wolf, Debut, Sanh, Chaumond, Delangue, Moi, Cistac, Rault, Louf, Funtowicz, Davison, Shleifer, von Platen, Ma, Jernite, Plu, Xu, Scao, Gugger, Drame, Lhoest, and Rush}]{wolf2020huggingfacestransformersstateoftheartnatural}
Thomas Wolf, Lysandre Debut, Victor Sanh, Julien Chaumond, Clement Delangue, Anthony Moi, Pierric Cistac, Tim Rault, Rémi Louf, Morgan Funtowicz, Joe Davison, Sam Shleifer, Patrick von Platen, Clara Ma, Yacine Jernite, Julien Plu, Canwen Xu, Teven~Le Scao, Sylvain Gugger, Mariama Drame, Quentin Lhoest, and Alexander~M. Rush. 2020.
\newblock \href {http://arxiv.org/abs/1910.03771} {Huggingface's transformers: State-of-the-art natural language processing}.

\bibitem[{Wu et~al.(2021{\natexlab{a}})Wu, Rao, Lan, Sun, and Qi}]{wu2021unified}
Lianwei Wu, Yuan Rao, Yuqian Lan, Ling Sun, and Zhaoyin Qi. 2021{\natexlab{a}}.
\newblock Unified dual-view cognitive model for interpretable claim verification.
\newblock In \emph{Proceedings of the 59th Annual Meeting of the Association for Computational Linguistics and the 11th International Joint Conference on Natural Language Processing (Volume 1: Long Papers)}, pages 59--68.

\bibitem[{Wu et~al.(2021{\natexlab{b}})Wu, Rao, Sun, and He}]{wu2021evidence}
Lianwei Wu, Yuan Rao, Ling Sun, and Wangbo He. 2021{\natexlab{b}}.
\newblock Evidence inference networks for interpretable claim verification.
\newblock In \emph{Proceedings of the AAAI Conference on Artificial Intelligence}, volume~35, pages 14058--14066.

\bibitem[{Wu et~al.(2013)Wu, Atkin, Lau, Lin, and Mou}]{Wu2013AgendaSA}
Yanfang Wu, David~J. Atkin, T.~Y. Lau, Carolyn~A. Lin, and Yikun Mou. 2013.
\newblock \href {https://api.semanticscholar.org/CorpusID:143206472} {Agenda setting and micro-blog use: An analysis of the relationship between sina weibo and newspaper agendas in china}.
\newblock \emph{Social media and society}, 2.

\bibitem[{Wu et~al.(2021{\natexlab{c}})Wu, Zhan, Zhang, Wang, and Xu}]{Wu2021}
Yang Wu, Pengwei Zhan, Yunjian Zhang, Liming Wang, and Zhen Xu. 2021{\natexlab{c}}.
\newblock Multimodal fusion with co-attention networks for fake news detection.
\newblock In \emph{Findings of the Association for Computational Linguistics: ACL-IJCNLP 2021}, pages 2560--2569.

\bibitem[{Yang et~al.(2018)Yang, Lin, Cohen, and Lee}]{Yang2018}
Zhicheng Yang, Tsung-Yi Lin, Scott Cohen, and Honglak Lee. 2018.
\newblock Graph r-cnn for scene graph generation.
\newblock In \emph{Proceedings of the European Conference on Computer Vision (ECCV)}.

\bibitem[{Yu et~al.(2017{\natexlab{a}})Yu, Liu, Wu, Wang, and Tan}]{ijcai2017p545}
Feng Yu, Qiang Liu, Shu Wu, Liang Wang, and Tieniu Tan. 2017{\natexlab{a}}.
\newblock \href {https://doi.org/10.24963/ijcai.2017/545} {A convolutional approach for misinformation identification}.
\newblock In \emph{Proceedings of the Twenty-Sixth International Joint Conference on Artificial Intelligence, {IJCAI-17}}, pages 3901--3907.

\bibitem[{Yu et~al.(2017{\natexlab{b}})Yu, Liu, Wu, Wang, Tan et~al.}]{yu2017convolutional}
Feng Yu, Qiang Liu, Shu Wu, Liang Wang, Tieniu Tan, et~al. 2017{\natexlab{b}}.
\newblock A convolutional approach for misinformation identification.
\newblock In \emph{IJCAI}, pages 3901--3907.

\bibitem[{Zannettou et~al.(2019)Zannettou, Sirivianos, Blackburn, and Kourtellis}]{Zannettou2019}
Savvas Zannettou, Michael Sirivianos, Jeremy Blackburn, and Nicolas Kourtellis. 2019.
\newblock The web of false information: Rumors, fake news, hoaxes, clickbait, and various other shenanigans.
\newblock \emph{Journal of Data and Information Quality (JDIQ)}, 11(3):1--37.

\bibitem[{Zhang et~al.(2017)Zhang, Lin, Goyal, Huang, Efros, and Darrell}]{Zhang2017}
Ji~Zhang Zhang, Peng-Tao Lin, Raghav Goyal, Joey Huang, Alexei~A Efros, and Trevor Darrell. 2017.
\newblock Graphical contrastive losses for scene graph parsing.
\newblock In \emph{Proceedings of the IEEE International Conference on Computer Vision (ICCV)}.

\bibitem[{Zhang et~al.(2021)Zhang, Cao, Li, Sheng, Zhong, and Shu}]{zhang2021mining}
Xueyao Zhang, Juan Cao, Xirong Li, Qiang Sheng, Lei Zhong, and Kai Shu. 2021.
\newblock Mining dual emotion for fake news detection.
\newblock In \emph{Proceedings of the Web Conference 2021}, pages 3465--3476.

\bibitem[{Zhao et~al.(2021)Zhao, Wang, Li, Yang, and Zhang}]{zhao2021semantic}
Boxin Zhao, Xindi Wang, Wenbin Li, Shuang Yang, and Haohong Zhang. 2021.
\newblock Semantic propositional image caption evaluation.
\newblock \emph{arXiv preprint arXiv:2101.02151}.

\bibitem[{Zhou et~al.(2020)Zhou, Wu, and Zafarani}]{Zhou2020}
Xinyi Zhou, Jindi Wu, and Reza Zafarani. 2020.
\newblock Safe: Similarity-aware multi-modal fake news detection.
\newblock In \emph{Pacific-Asia Conference on Knowledge Discovery and Data Mining}, pages 354--367. Springer.

\bibitem[{Zhou and Zafarani(2020)}]{ZhouZafarani2020}
Xinyi Zhou and Reza Zafarani. 2020.
\newblock \href {https://arxiv.org/abs/1812.00315} {A survey of fake news: Fundamental theories, detection methods, and opportunities}.
\newblock \emph{ACM Computing Surveys (CSUR)}, 53(5):1--40.

\bibitem[{Zhou et~al.(2023)Zhou, Yang, Ying, Qian, and Zhang}]{Zhou2023}
Yangming Zhou, Yuzhou Yang, Qichao Ying, Zhenxing Qian, and Xinpeng Zhang. 2023.
\newblock \href {https://doi.org/10.1145/3591106.3592271} {Multi-modal fake news detection on social media via multi-grained information fusion}.
\newblock In \emph{Proceedings of the 2023 ACM International Conference on Multimedia Retrieval (ICMR '23)}, pages 343--352, New York, NY, USA. Association for Computing Machinery.

\end{thebibliography}

\appendix
\section*{Appendix}
\section{Experiment Details}\label{expAppendix}
\subsection*{A.1 Optimizer and Hyperparameters}
For training the model, we used the \textbf{Adam optimizer}. Table \ref{hyper} summarizes the optimizer and key hyperparameters:

\begin{table}[h!]
    \centering
    \begin{tabular}{|l|c|}
        \hline
        \textbf{Hyperparameter} & \textbf{Value} \\
        \hline
        Optimizer               & Adam \\
        Learning Rate           & 1e-5 \\
        Weight Decay            & 1e-7 \\
        Dropout Rate            & 0.3 \\
        Beta1 (Adam parameter)  & 0.9 \\
        Beta2 (Adam parameter)  & 0.999 \\
        Epsilon (Adam parameter) & 1e-8 \\
        \hline
    \end{tabular}

    \caption{Optimizer and Hyperparameter Details}
    \label{hyper}
\end{table}

We conducted a hyperparameter search for optimal learning rate, weight decay, and dropout values.

\subsection*{A.2 Specific Versions of Pre-Trained Modules}
In our implementation, we leveraged several pre-trained modules to extract features from text and images. The versions of these pre-trained modules are as follows:
\begin{itemize}
    \item \textbf{BERT Model:} The text embeddings are generated using \textbf{BERT-base-uncased} \cite{Devlin2019}, with 12 layers, 768 hidden units, 12 attention heads, and 110M parameters.
    We use Hugging Face Transformers library \cite{wolf2020huggingfacestransformersstateoftheartnatural} to load and utilise BERT.
    
    \item \textbf{TSG \& VSG Generators:} For generating \textbf{textual scene graphs (TSG)}, we utilized a pre-trained \textbf{TSG model} based on \textbf{Spice} \cite{spice2016}. The \textbf{visual scene graph (VSG)} model is derived from the \textbf{ReITR framework} \cite{cong2023reltr}.
\end{itemize}

\subsection*{A.3 Hardware Details}
We conducted all experiments on Nvidia 1080 Ti 12 GB GPUs, mostly using a single GPU. Batch sizes ranged from 8 to 32, depending on the dataset.

\end{document}